\documentclass{article}

\PassOptionsToPackage{numbers, compress}{natbib}

\usepackage[preprint]{neurips_2020}

\usepackage[utf8]{inputenc} 
\usepackage[T1]{fontenc}    
\usepackage{hyperref}       
\usepackage{url}            
\usepackage{booktabs}       
\usepackage{amsfonts}  
\usepackage{amsmath}
\usepackage{nicefrac}       
\usepackage{microtype}      
\usepackage{multirow}
\usepackage[inkscapearea=page]{svg}

\title{Dialogue-Contextualized Re-ranking for Medical History-Taking}

\author{Jian Zhu\thanks{Work done during an internship at Curai Health} \\
     University of British Columbia \\
  \texttt{jian.zhu@ubc.ca}\\
  \AND
  Ilya Valmianski \thanks{Work done while at Curai Health} \\
   AuxHealth \\
  \texttt{ilya@auxhealth.io} \\
  \And
  Anitha Kannan \\
   Curai Health \\
  \texttt{anitha@curai.com} \\
  }

\begin{document}

\maketitle

\begin{abstract}

AI-driven medical history-taking is an important component in symptom checking, automated patient intake, triage, and other AI virtual care applications. As history-taking is extremely varied, machine learning models require a significant amount of data to train. To overcome this challenge, existing systems are developed using indirect data or expert knowledge. This leads to a training-inference gap as models are trained on different kinds of data than what they observe at inference time. In this work, we present a two-stage re-ranking approach that helps close the training-inference gap by re-ranking the first-stage question candidates using a dialogue-contextualized model. For this, we propose a new model, global re-ranker, which cross-encodes the dialogue with all questions simultaneously, and compare it with several existing neural baselines. We test both transformer and S4-based language model backbones. We find that relative to the expert system, the best performance is achieved by our proposed global re-ranker with a transformer backbone, resulting in a 30\% higher normalized discount cumulative gain (nDCG) and a 77\% higher mean average precision (mAP). As part of this work, we also release pre-trained checkpoints for bi-directional and autoregressive S4 models trained on Wikipedia and PubMed data. 
\end{abstract}

\section{Introduction}
\label{sec:introduction}

History taking is a critical component of a medical encounter \citep{hampton1975relative}. It involves collecting relevant patient-reported information such as   presenting symptoms, patient concerns as well as the past medical, psychological and social history. This information forms the basis of subsequent patient triage, diagnosis, and care planning.  While history taking is an important component of the medical encounter, it is also one of the most time-consuming components \citep{Soltau2021,Chen2020} and when done incompletely can lead to triage, diagnostic, and treatment errors \citep{hampton1975relative}. 



Creating tools for automating portions of history taking has been an on-going effort for more than five decades \citep{watson1965health}. The simplest of such tools are static pre-visit questionnaires that are now used widely in US healthcare. However, static questionnaires tend to be long, ask not very relevant questions, and are not customized to patients' needs.  More recently, there has been work on building intelligent systems that can adjust questions based on patient responses (see \citep{pmlr-v158-valmianski21a, compton2021medcod} and citations therein). However, developing the medical reasoning necessary for these systems is difficult. Existing approaches include using reinforcement learning with simulated patients \citep{tang2016inquire,kao2018context}, supervised learning on clinical notes \citep{pmlr-v158-valmianski21a}, and expert systems \citep{compton2021medcod}.


In all of the previous works, the medical reasoning system was built on top of data that was a proxy for real doctor-patient interactions. This is because, on the one hand, there is little available data consisting of doctor-patient history-taking dialogue, on the other hand, the space of possible questions asked during history-taking is very large. Thus, training a history-taking model requires significant amounts of labeled interaction data. However, this data is more readily available from indirect sources such as medical notes, expert knowledge, or simulations. This creates a training-inference gap: the data that is used to train the model is not fully representative of the data that the model sees at inference time. 

 

This training-inference gap has a significant impact on the quality of history taking, especially since only a few questions can be asked in a given encounter. This calls for an approach that reconciles the difficulty of supporting a large set of potential questions to a small set of pertinent questions attuned to the patient's health issue, with only a small amount of direct training data.


In this paper, we start with an expert system and show how to use a relatively small amount of real doctor-patient dialogue data to close this training-inference gap. We take inspiration from the information retrieval literature where ``retrieve and re-rank'' is a popular paradigm for computationally efficient retrieval of documents from a large corpus \citep{nogueira2019passage,lin2021pretrained}. In our case, the ``retrieve'' part is performed by the expert systems which retrieves a list of possible questions to ask the patient, and a dialogue-trained re-ranker then ``re-ranks'' the possible questions. Because the re-ranking model takes the original expert system's candidate questions, it does not need to predict over the space of all possible questions. Instead, it only needs to re-rank from a much smaller subset, which greatly simplifies the machine-learning task. Our model takes both the previous dialogue and the possible questions as free text entries, which means that the system can operate even if the underlying expert system is replaced with something else. 

Our contributions are as follows:

\begin{enumerate}
    \item We propose a two-step approach to history-taking question selection where we use an expert system to retrieve a list of candidate questions and then use a machine-learned re-ranker to get the top question to ask.
    \item We propose a  novel ``global re-ranker'' which embeds both the preceding dialogue and candidate  questions into a single long string. We then train long context language models to predict the relevance of each question simultaneously.
    \item We perform a careful study of other re-rankers for this task.  This includes different architectures such as bi-encoder, cross-encoder, and autoregressive re-rankers. We examine different long context models including S4 (bi-directional and autoregressive), Nystromformer (bi-directional, variants with Nystrom attention and with full attention), and LongT5 (autoregressive). We examine the effect of different loss functions from the pointwise, pairwise, and listwise families. Finally, we perform some ablation studies on the context length and the initial retrieval ordering.
    \item We release checkpoints for S4 pre-trained both bidirectionally and autoregressively on the English subset of Wikipedia\footnote{\url{https://huggingface.co/datasets/wikipedia}} and Pubmed PMC Open Access Subset\footnote{\url{https://www.ncbi.nlm.nih.gov/pmc/tools/textmining/}} datasets.
    \item We find that our global re-ranker approach performs better than other more traditional approaches. Furthermore, all re-rankers significantly improve the original expert system performance. We also find that S4-based, while worse than full-attention transformers, is competitive with the Nystrom-attention transformer.
\end{enumerate}

\section{Generalizable Insights about Machine Learning in the Context of Healthcare}
One of the main challenges in using deep learning for healthcare is the lack of large annotated datasets. Obtaining large amounts of annotated data is  costly and time-consuming because annotations need to be provided by trained healthcare professionals. Recent works have successfully leveraged the progress in the development of large language models that are trained on web-scale data. In many tasks, including medical history taking discussed in this paper, this approach introduces a training-inference gap: the data used to
train the model does not fully represent the data that the model sees at inference time.  In this context, our approach of retrieving a candidate set of answers (we use an expert system as the base model to provide candidates, but this can come from a large language model, too) and then using a learned reranker based on small amounts of labeled data is a promising alternative. As we show in this paper, such a reranker can be trained from public data sources and then fined tuned to the task.

\section{Related work}
\label{sec:related-work}

We study re-ranking history-taking recommendations based on doctor-patient dialogue. These dialogues tend to be long and exceed the typical token-length limits of transformer models. As such, there are two bodies of literature relevant to this work. \S~\ref{sec:rel-transformers} discusses work on modern neural long-range language models that are able to encode the entire doctor-patient conversation. \S~\ref{sec:rel-reranking} discusses work on re-ranking algorithms that can take the encoded dialogue and use it to re-rank history-taking questions.

\subsection{Long-range transformers}
\label{sec:rel-transformers}
Transformers \citep{vaswani2017attention} have become the mainstream architecture for natural language processing. With the self-attention mechanism, transformers can attend to all tokens in a sequence simultaneously, thereby being more powerful than classical architectures like convolutional \citep{lecunzipcodes} or long short-term memory (LSTM) \citep{hochreiter1997long} networks. However, due to its $O(n^2)$ complexity, the original transformer cannot process long sequences efficiently. Popular pre-trained language models such as BERT \citep{devlin-etal-2019-bert} and RoBERTa \citep{liu2019roberta} can only process up to 512 tokens. Efforts have been made to reduce the computational complexity of transformers, as a variety of efficient transformers that can process long text sequences have been proposed, such as Reformer \citep{kitaev2019reformer}, Linformer \citep{wang2020linformer}, Longformer \citep{beltagy2020longformer}, BigBird \citep{zaheer2020big}, Performer \citep{choromanski2020rethinking}, Nystromformer \citep{xiong2021nystromformer}, LongT5 \citep{guo2021longt5}, etc. In addition to transformers, alternative approaches have also been shown to be promising for processing very long sequences, notably state-space models \citep{gu2021efficiently,gu2022parameterization,mehta2022long}. The Structured State Space Sequence model (S4) \citep{gu2021efficiently} has significantly outperformed many long-range transformers in the Long Range Arena benchmark \citep{tay2020long}. In this paper we utilize the Nystromformer\citep{xiong2021nystromformer} and the S4 model\citep{gu2021efficiently} as the possible non-autoregressive backbones and LongT5 as  an autoregressive backbone.

\subsection{Re-ranking}
\label{sec:rel-reranking}
Modern information retrieval (IR) or question answering (QA) systems are usually divided into two stages \citep{nogueira2019passage,lin2021pretrained}. In the first stage,  given a query, a large number of documents are retrieved using an efficient method. In the second stage, a computationally intensive but more accurate method is used to re-rank documents retrieved in the first stage. This is analogous to our problem statement where we use an expert system to retrieve a set of relevant history-taking questions, and then use a re-ranking algorithm as the second stage. Relevant related work addresses both the architectural choices that can be made for the re-rankers, as well as the loss functions used to train them.

\paragraph{Architectures for second-stage re-ranking.} There are three main type of architectures (1) bi-encoders \citep{lin2021pretrained,nogueira2019multi,reimers-gurevych-2019-sentence,thakur2021augmented} (2) cross-encoders \citep{lin2021pretrained,nogueira2019multi,humeau2019poly} and (3) autoregressive re-rankers \citep{nogueira-etal-2020-document,pradeep2021expando,min-etal-2021-joint}. In the bi-encoder architecture, the query and the candidate document are encoded into vector representations by two separate encoders (though these two encoders can share the same weights \citep{reimers-gurevych-2019-sentence}). The relevance score is calculated as the distance between the two vector representations. In a cross-encoder, a query and a candidate document are concatenated together and fed into the cross-encoder in a single pass. In most use cases, cross-encoders outperform bi-encoders in document retrieval and ranking \citep{thakur2021augmented,humeau2019poly}; however, bi-encoders are usually more efficient than cross-encoders, as all documents in a given corpus can be pre-computed and stored as dense embeddings for retrieval, thereby avoiding repeated computations \citep{reimers-gurevych-2019-sentence}. Recent sequence-to-sequence models such as T5 \citep{raffel2020exploring} have also been applied to autoregressive re-ranking. In this approach, the query and the documents are usually encoded by the encoder and the decoder either predicts whether the document is relevant \citep{nogueira-etal-2020-document} or directly generates the retrieved text in response to the query \citep{min-etal-2021-joint}. 

\paragraph{Scoring functions for ranking.} There are also several scoring paradigms possible for ranking: (1) ``pointwise'' scores where the relevance of a query is computed on a per-document basis (similar to cross-encoders)  \citep{nogueira2019multi}, (2) ``pairwise'' scores where documents are ranked relative to each other in pairs \citep{nogueira2019multi}, and (3) ``listwise'' scores where a list of candidates are ranked simultaneously \citep{liu2009learning}. Prior studies show that pairwise and listwise approaches tend to outperform pointwise approaches \citep{nogueira2019multi,pradeep2021expando,min-etal-2021-joint,zerveas2021coder,chen2021co}. In this study, we compare all three (`pointwise', `pairwise', and `listwise') approaches for re-ranking history-taking questions but mainly focus on listwise approaches.

\begin{figure}[!ht]
    \centering
    \includegraphics[width=\textwidth]{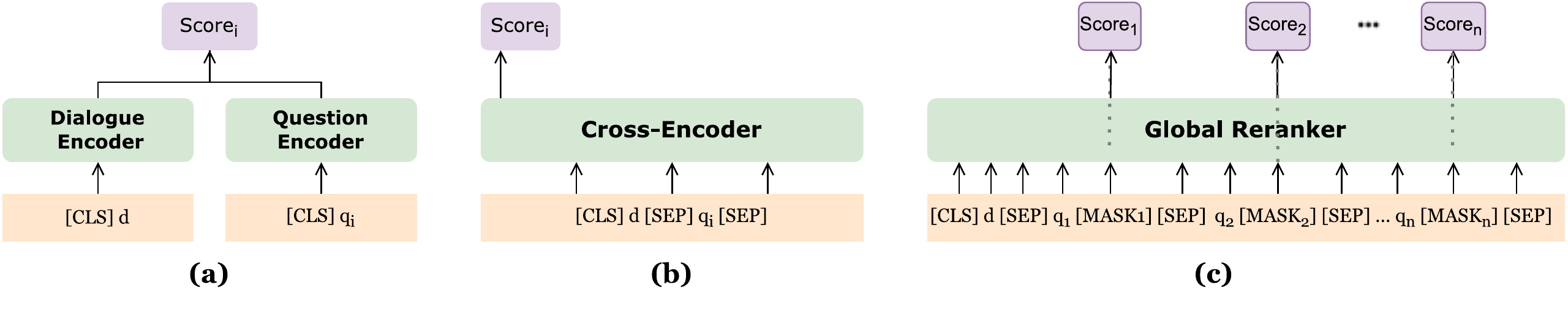}
    \caption{(a) Bi-encoder, (b) the cross-encoder  and (c) the proposed global reranker. }
    \label{fig:models}
    \vspace{-1em}
\end{figure}

\paragraph{Connection to our proposed global re-ranker.} In this paper we propose a novel re-ranker approach we call ``global re-ranker.'' In this approach, all candidate documents are concatenated into a single input that is then processed by a long context language model. For schematic comparison between bi-encoder, cross-encoder, and global re-ranker please see \autoref{fig:models}. For a more detailed description of the method see \S~\ref{sec:global-re-ranker}. Concatenating pairs of documents into a single string has been previously done both in bi-directional \citep{nogueira2019multi} and autoregressive \citep{pradeep2021expando} paradigms. These models only investigate pairwise scoring due to length constraints imposed by the pre-trained transformer. For listwise ranking with more than two documents, previous approaches focused on ranking only the extracted embeddings \citep{chen2021co,ai2019learning}, which doesn't model the deep semantic relationships between candidate documents.

\section{Closing the train-inference gap with re-ranking}
\label{sec:method}

\begin{figure}[!ht]
    \centering
    \includegraphics[width=0.8\textwidth]{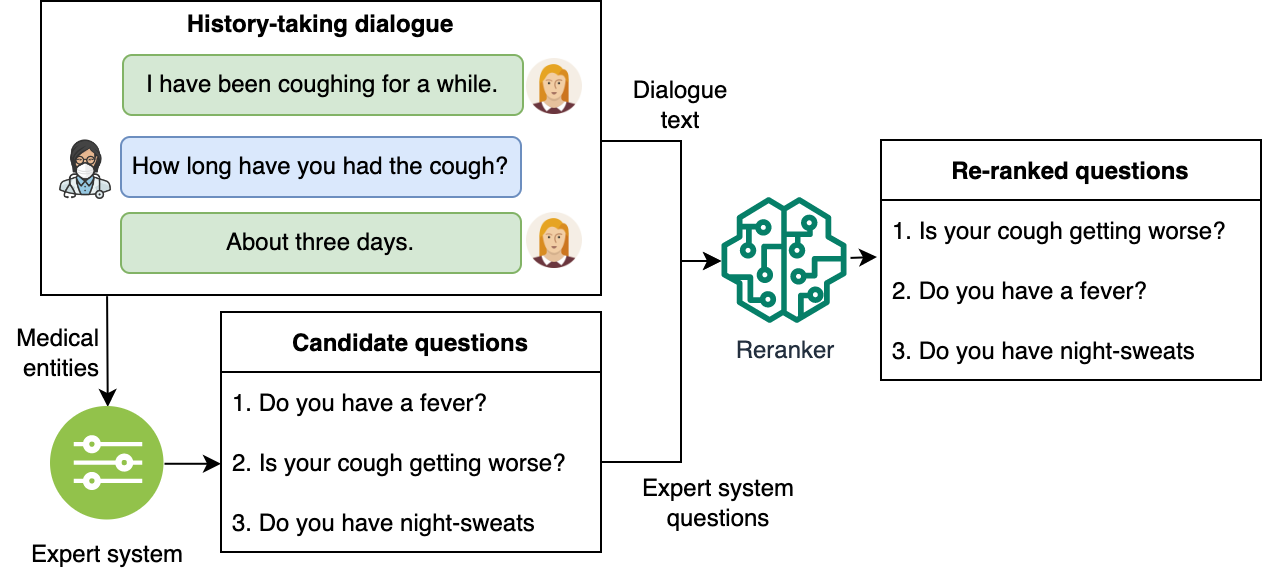}
    \caption{Overview of the proposed two-stage history-taking workflow. An expert system suggests candidate questions based on relevant entities extracted from the dialogue. A machine-learned deep neural network re-ranker then re-ranks the candidate questions based on the dialogue text.}
    \label{fig:flow}
    \vspace{-1em}
\end{figure}

An overview of our approach to closing the train-inference gap in an existing history-taking system can be seen in \autoref{fig:flow}. We first use an expert system to suggest relevant history-taking questions and then use a deep neural network contextualized by the entire doctor-patient dialogue to re-rank expert system suggestions.

The goal of re-ranking is,  given the prior dialogue context $\boldsymbol{d}$ and a list of $n$ candidate history-taking questions $Q=[\boldsymbol{q}_1, \boldsymbol{q}_2, \dots, \boldsymbol{q}_n]$, to generate a new list $Q'$ which consists of (possibly reordered) questions from $Q$ such that the higher relevance questions appear earlier in the sequence. In our case, the candidate questions are generated using an in-house Expert System, and the ground truth labels $\boldsymbol{y}=[y_1,y_2,\dots,y_n], y_i\in\{0,1\}$ represent whether a doctor~asked a given recommended question ($1$ if the question was asked, $0$ if the question was not asked). A doctor may ask multiple questions at the same time, thus multiple elements of $\boldsymbol{y}$ can have a value of $1$, see \S~\ref{sec:data} for more details on how the ground truth is produced. Finally, in all of the models studied in this work, the re-ranking is achieved by assigning scores $\boldsymbol{s}=[s_1,s_2,\dots,s_n]$ to each question in $Q$, and then constructing $Q'$ by reordering using scores in $\boldsymbol{s}$.


\section{Global re-ranker}
\label{sec:global-re-ranker}

We propose the global re-ranker, an accurate and efficient listwise re-ranking method. 
In this approach (see \autoref{fig:models}(c) for a schematic), the history-taking dialogue and all candidate history-taking questions are concatenated into a single text input, using which the model then assigns the ranking scores to all questions simultaneously. The global re-ranker directly encodes all texts through the language model, thereby ensuring deep semantic interactions not only between the dialogue and the candidate questions but also between all candidate questions.

The input text to the global re-ranker is the concatenation of both the dialogue context and all the candidate questions: \texttt{[CLS]} $\boldsymbol{d}$ \texttt{[SEP]} $\boldsymbol{q}_1$ \texttt{[MASK\textsubscript{1}][SEP]} $\boldsymbol{q}_2$ \texttt{[MASK\textsubscript{2}]}\texttt{[SEP]} \dots  $\boldsymbol{q}_n$ \texttt{[MASK\textsubscript{n}][SEP]}, 
where the \texttt{[SEP]} token is used to mark the boundaries of candidate questions. The \texttt{[MASK\textsubscript{i}]} token is the pooling token for the preceding question $\boldsymbol{q}_i$. For each pooling token \texttt{[MASK\textsubscript{i}]}, the global reranker predicts a score $s_i$, which represents the relevance for $\boldsymbol{q}_i$. We also added type embeddings to every input tokens to indicate whether it belongs to the dialogue or the candidate questions. The actual number of candidate questions provided by the expert system ranged from 3 to 40.

While self-attention itself does not assume any inherent order of the input sequence, pretrained transformer models usually encode the text sequentially due to the presence of positional embeddings. In the current task, it is expected that a language model learns the sequential relations between words within $\boldsymbol{d}$ and $\boldsymbol{q}_i$. From our ablation experiments (see \S~\ref{sec:ablations}), we found that the best performance is achieved when the model is agnostic to the order of input questions $[\boldsymbol{q}_1, \boldsymbol{q}_2, \dots, \boldsymbol{q}_n]$. In order to remove the positional bias, we reset the positional embedding when each new question starts. 
 
We selected three different neural architectures to implement the global ranker, all of which can process long textual sequences. The first two approaches are based on the Nystromformer \citep{xiong2021nystromformer}, which was originally proposed to be an efficient transformer. We experiment with Nystromformer with both Nystrom attention turned on and turned off (in which case it uses full attention). We use Nystromformer as the base of our ``full attention'' transformer because this enables us to leverage the pretrained Nystromformer checkpoints that had been trained on long texts and retain the good performance of full attention. We learned from pilot experiments that other efficient transformers such as Longformer \citep{beltagy2020longformer} failed to converge. The third neural architecture is a state-space model, S4, which has been shown to process long sequences more effectively than many transformers \citep{gu2021efficiently}.

To train the global re-ranker, we compared a variety of loss functions across point-wise, pair-wise and listwise approaches in the learning-to-rank framework \citep{liu2009learning}. The point-wise baseline was trained with binary cross-entropy. For pairwise loss functions, we tested the RankNet \citep{burges2005learning} and LambdaRank \citep{burges2006learning}. The listwise loss functions we used were ListNet \citep{cao2007learning}, ListMLE \citep{xia2008listwise}, ApproxNDCG \citep{qin2010general} and NeuralNDCG \citep{Pobrotyn2021NeuralNDCG}, the latter two of which directly optimized the Normalized Discounted Cumulative Gain (NDCG) metrics. 

\section{Experiments}

\subsection{Data}
\label{sec:data}
The medical dialogue data was collected from a portion of real doctor-patient interactions collected on our text-based medical service platform. In a typical interaction, the physician asks a series of history-taking questions that can be entered either as free text or selected from a list of recommendations. These recommendations are made using the Expert System that forms the first stage in our proposed workflow. At each dialogue turn where recommended questions are asked, the doctor~selected questions are marked as relevant and the not-selected questions are marked as irrelevant. This forms a natural dataset of doctor~annotated selections on which we train our re-rankers.

The dataset consists of 13071 encounters. We filtered non-history-taking dialogue turns using in-house dialogue segmentation model, similar to \cite{wang2022learning}. The detailed statistics of our data are displayed in Table~\ref{tab:stats}. 


\begin{table}[t!]
\centering

\begin{tabular}{lccc}
\toprule
& \hspace{1em}\textbf{Train}\hspace{1em} & \hspace{1em}\textbf{Dev}\hspace{1em} & \hspace{1em}\textbf{Test}\hspace{1em}
\\ \midrule
\textbf{Num. Encounters} & 12105 & 311 & 655 \\
\textbf{Num. Samples} & 26106 & 626 & 1361 \\
\textbf{Avg. Length of Dialog.} & 287.2 & 374.7 & 288.8 \\
\textbf{Num. Selected Questions} & 4.0 & 3.8 & 4.1 \\
\textbf{Num. Candidate Questions} & 27.9 & 26.6 & 28.1 \\
\textbf{Avg. Length of Questions} & 8.0 & 8.1 & 8.0 \\ \bottomrule
\end{tabular}
\vspace{0.5em}
\caption{Statistics of different data splits. Text lengths were calculated based on words. }
\label{tab:stats}
\end{table}

\subsection{Metrics} 

For evaluation, we adopted two common ranking metrics, normalized discounted cumulative gain (nDCG)\citep{nDCGmetric} and mean average precision (mAP) \citep{lin2021pretrained}. The mAP assumes binary relevance whereas nDCG can work with both binary and continuous relevance. Specifically for global re-rankers, the average metrics over 5 repeated runs of evaluations were reported. In each run, the order of candidate questions fed to the global re-ranker was randomly reshuffled to mitigate positional biases.

\subsection{Baseline approaches} 
In addition to the global ranker, we also implement three widely adopted baseline ranking approaches: bi-encoder, cross-encoder, and autoregressive re-ranker. 

\paragraph{Bi-encoder.} In the bi-encoder architecture (see Figure~\ref{fig:models}(a)), the dialogue query and the candidate questions are encoded by two separate encoders $f_D$ and $f_Q$, and the relevance score between the two resulting vector representations are computed with cosine similarity. The bi-encoder learns an embedding space where the dialogue representation is close to the most relevant questions while being distant from less relevant questions. The training objective is to minimize the InfoNCE loss function \citep{oord2018representation} through contrastive learning with 7 negatives randomly sampled from the list of recommended candidate questions by the Expert System. The temperature parameter of the InfoNCE loss was set to 0.05 throughout the training \citep{gao-etal-2021-simcse}. 

\paragraph{Cross-encoder.} In the cross-encoder architecture (see Figure~\ref{fig:models}(b)) the prior dialogue is concatenated with a candidate question. The cross-encoder $f_C$ assigns a relevance score to this candidate question using a classification head on top of the contextual representation of the dialogue and the query. We consider transformers and S4-based models. For transformers, the \texttt{[CLS]} token is treated as the contextual representation. For the bi-directional S4 re-rankers, we use average pooling of the last layer to obtain the contextual representations. All cross-encoder variants are trained with the binary cross-entropy loss. 

\paragraph{Autoregressive re-ranker.} We also consider autoregressive re-rankers \citep{nogueira-etal-2020-document,pradeep2021expando}. For a transformer baseline, we use a pre-trained LongT5 \citep{guo2021longt5}. The query and the document are concatenated together to form the input sequence: \texttt{Query:} $d$ \texttt{Document:} $q_i$ \texttt{Relevant:}, which is fed into the encoder. The decoder then predicts \texttt{true} for relevant documents or \texttt{false} for irrelevant documents. During inference, a softmax function is applied to the logits of the \texttt{true} and the \texttt{false} tokens to normalize the results across multiple queries. 

For autoregressive S4, when we followed the Long-T5 method, we found it to highly unstable and failed to converge, similar to what was found in the literature \citep{nogueira-etal-2020-document} on its dependency certain keywords e.g., true/false.  Therefore, we followed the same setting as in the cross-encoder, except that the underlying model is autoregressive rather than bi-directional. Here, the concatenated dialogue and a candidate question are fed into the S4 re-ranker and the average pooling of the last layer is classified as either relevant or irrelevant through a classification head.


\subsection{Implementation} 

\paragraph{S4 model pretraining.} The S4 model was based on the original implementation of S4 language model\citep{gu2021efficiently}, in which the S4 layers were used as a drop-in replacement for the self-attention layers in a typical transformer. We implemented a 12-layer bidirectional and autoregressive S4 models.
We set the hidden dimensions to 768 in order to match the parameter count of mainstream pretrained transformers (such as BERT-base \citep{devlin-etal-2019-bert}), and the number of state-space machines (SSM) to 128 with 64 states for each SSM. 

Both bidirectional S4 and autoregressive S4 models were pretrained on large-scale texts. The autoregressive S4 was pretrained with the casual language modeling task on the whole English subset of Wikipedia. The second iteration of pretraining, initialized with the pretrained Wikipedia checkpoint, was on the whole Pubmed PMC Open Access Subset.  The bidirectional S4 models were pretrained on the same datasets but with the mask language modeling task using the same masking settings as in BERT \citep{devlin-etal-2019-bert}. The maximum sequence length for pretraining was set to 8192 and the effective batch size was 256. 

All models were optimized with AdamW optimizer with a learning rate of 1e-4 and the learning rate was dynamically adjusted using the Cosine Scheduler with a warm-up step of 1000. The pretraining took place on 8$\times$RTX 3090 GPU with 24GB of memory. The training was stopped when the evaluation loss stopped to decrease ($\sim$12k steps for all models). The autoregressive and bi-direction checkpoints pre-trained on these datasets will be released together with this paper.

\paragraph{Transformer implementation.} Transformer models were all implemented through the \texttt{Transformers} package \citep{wolf2020transformers} with default dimensions. The autoregressive model was LongT5 \citep{guo2021longt5} initialized by the \texttt{long-t5-tglobal-base} checkpoint. Other transformers were based on the Nystromformer \citep{xiong2021nystromformer} with initialization from the public checkpoint \texttt{uw-madison/nystromformer-4096}.

\paragraph{Re-ranker training.} For global re-rankers, the maximum input length was set to 4096 with an effective batch size of 32. For other models, the effective batch size was 64 and the maximum length was 2048, as this length was enough to cover almost all of the data samples. Models were trained with a maximum of 5 epochs and only the model with the best validation performance was kept. 
All models were trained using the AdamW optimizer \citep{loshchilov2018decoupled} with a learning rate of 5e-5. We used a cosine scheduler with a warm-up step of 1000 to automatically adjust the learning rate during training. All ranking models were trained on a single V100 GPU with 16GB of memory.

\section{Results}
\label{sec:results}

\subsection{Main results}

\begin{table}[ht]
\centering
\begin{tabular}{lllll}
\toprule
\multicolumn{1}{c}{\multirow{2}{*}{\textbf{Model}}} &
  \multicolumn{2}{c}{\textbf{Dev}} &
  \multicolumn{2}{c}{\textbf{Test}} \\ \cline{2-5} 
\multicolumn{1}{l}{} &
  \multicolumn{1}{l}{\textbf{nDCG}} &
  \multicolumn{1}{l}{\textbf{mAP}} &
  \multicolumn{1}{l}{\textbf{nDCG}} &
  \textbf{mAP} \\ \midrule
\textbf{Expert System} (Baseline) & 0.592 & 0.383 & 0.570 & 0.349 \\\midrule
\textbf{Bi-encoder} (Transformer) & 0.690 & 0.548 & 0.677 & 0.531  \\\midrule
\textbf{Cross-encoder}\\\midrule
\hspace{0.2in}Transformer & 0.718 & 0.584 & 0.706 & 0.566 \\
\hspace{0.2in}Nystromformer & 0.653 & 0.496  & 0.654 & 0.497 \\
\hspace{0.2in}Bidirectional S4 (Wiki Pretraining) & 0.648 & 0.490 & 0.641 & 0.481 \\
\hspace{0.2in}Bidirectional S4 (Pubmed Pretraining) & 0.643 & 0.483 & 0.630 & 0.464   \\\midrule
\textbf{Autoregressive Re-ranker} \\\midrule
\hspace{0.2in} LongT5-base & 0.690 & 0.546 & 0.678 & 0.529\\
\hspace{0.2in} Autoregressive S4 (Wiki Pretraining) & 0.658 & 0.502 & 0.648 & 0.490 \\
\hspace{0.2in} Autoregressive S4 (Pubmed Pretraining) & 0.654 & 0.498 & 0.642 & 0.484\\
\midrule
\multicolumn{5}{l}{\textbf{Global Re-ranker}} \\\midrule
\hspace{0.2in} Transformer \\
\hspace{0.5in} + Pointwise loss: BCE & \textbf{0.744} & \textbf{0.618} & \textbf{0.743} & \textbf{0.618} \\
\hspace{0.5in} + Pairwise loss: RankNet & 0.739 & 0.612 & 0.735 & 0.603\\
\hspace{0.5in} + Pairwise loss: LambdaLoss & 0.739 & 0
616 & 0.739 & 0.612 \\
\hspace{0.5in} + Listwise loss: ListNet & 0.737 & 0.609 & 0.740 & 0.610 \\
\hspace{0.5in} + Listwise loss: ListMLE  & 0.727 & 0.597 &  0.721 &0.587  \\
\hspace{0.5in} + Listwise loss: ApproxNDCG & 0.701 & 0.555 & 0.697 & 0.550 \\
\hspace{0.5in} + Listwise loss: NeuralNDCG & 0.742 & 0.617 & 0.741 & 0.612 \\
\hspace{0.2in} Nystromformer \\
\hspace{0.5in} + Pointwise loss: BCE & 0.684 & 0.537 & 0.678 & 0.530 \\
\hspace{0.2in} Bidirectional S4 (Wiki Pretraining) \\
\hspace{0.5in} + Pointwise loss: BCE & 0.667 & 0.516 & 0.663 & 0.510 \\
\hspace{0.2in}  Bidirectional S4 (PubMed Pretraining) \\
\hspace{0.5in} + Pointwise loss: BCE & 0.697 & 0.556 & 0.670 & 0.518 \\\bottomrule
\end{tabular}
\vspace{0.5em}
\caption{Results of reranking experiments.}
\label{tab:main-result}
\end{table}


Our main results are summarized in Table~\ref{tab:main-result}. All neural re-ranking models outperform the baseline Expert System in both metrics, suggesting that re-ranking does up-rank the more relevant history-taking questions. Among the neural baselines, the transformer-based cross-encoder outperforms the bi-encoder, which is consistent with previous findings \citep{nogueira2019multi}. Surprisingly, the LongT5 autoregressive re-ranker, despite having more parameters (220M parameters), also performs worse than the cross-encoder ($\sim$110M parameters).

The best performance is achieved by the global re-ranker for both transformer and S4 architectures, regardless of the loss functions chosen. Among the various loss functions, the pointwise binary cross-entropy (BCE) performs the best. Our hypothesis is that since our ground truth relevance scores are binary rather than continuous, the current task does not make full use of the listwise loss functions. 

The effectiveness of the global re-ranker lies in the fact that it attends to the semantic interactions not only between the dialogue and the candidate questions but also between the candidate questions themselves. This allows the model to exploit the dependencies between history-taking questions, such as co-occurrence statistics, to improve ranking outcomes.  

It is also worth noting that, despite its outstanding performance in some long sequence processing benchmark \citep{gu2021efficiently}, S4 still lags behind transformers in the current task. One reason could be that the S4 model here has only been pre-trained on a comparatively small amount of texts, while transformers have been pre-trained on huge amounts of texts. Furthermore, the text sequences in our task range from a few hundred to about three thousand words, which might not be long enough for S4 to reveal its full potential.

\subsection{Ablation analysis}
\label{sec:ablations}
We conducted ablation analyses on the global re-ranker to assess the impact of dialogue context length, the effect of type embeddings, and the effect of shuffling candidate question order. The results are displayed in Table~\ref{tab:ablation}. 

\paragraph{Context length ablations.} When ablating on context length, only the \textit{last} $N$ tokens of the dialogue were considered (full model is 4096 tokens, ablation are 3072, 2048, and 1024 tokens). While most of the text sequences were shorter than 1000 tokens, truncating texts still decreases test performance on some text sequences that are particularly long (longer than 1024), as some important information could be removed. In general, the global re-ranker benefits from getting more dialogue contexts, though this benefit seems to diminish after expanding to more than 2048 tokens. 

\paragraph{Effect of position and type embeddings.} We find that the removal of type embeddings (which are learned embeddings that differentiate whether the token is from dialogue or a candidate question) has almost no impact on the test performance. We reset the positional embeddings for each candidate questions in the input sequence, as this might help the model learn to be agnostic to the order of questions. We trained a model that used sequential positional embeddings for the input sequence. It turned out that positional embeddings played a minor role in training the global re-ranker.

\paragraph{Effect of shuffling.} We tested the importance of permutation invariance with regard to the order of input candidate questions. The list of candidate questions $[\boldsymbol{q}_1, \boldsymbol{q}_2, \dots, \boldsymbol{q}_n]$ were concatenated with the prior dialogue as an input to the model. We found that while the expert system should produce questions in order or relevance, performance was significantly higher when the model was trained with shuffled order. We believe that this forces the model to learn to re-rank the questions without falling back to the original order of the candidate questions.

\begin{table}[ht]
\centering
\begin{tabular}{lllll}
\toprule
\multicolumn{1}{c}{\multirow{2}{*}{\textbf{Ablation}}} &
  \multicolumn{2}{c}{\textbf{Dev}} &
  \multicolumn{2}{c}{\textbf{Test}} \\ \cline{2-5} 
\multicolumn{1}{l}{} &
  \multicolumn{1}{l}{\textbf{nDCG}} &
  \multicolumn{1}{l}{\textbf{mAP}} &
  \multicolumn{1}{l}{\textbf{nDCG}} &
  \textbf{mAP} \\ \midrule
\hspace{0.2in}  - Maximum length: 4096 (Full) & 0.744 & 0.618 & 0.743 & 0.618 \\
\hspace{0.2in}  - Maximum length: 3072 & 0.741 & 0.614 & 0.739 & 0.611 \\
\hspace{0.2in}  - Maximum length: 2048 & 0.746 & 0.622 & 0.747 & 0.622 \\
\hspace{0.2in}  - Maximum length: 1024 & 0.747 & 0.623 & 0.737 & 0.609  \\
\hspace{0.2in} - No type embedding & 0.733 & 0.610  & 0.732 & 0.607\\
\hspace{0.2in} - Sequential position embedding & 0.749 & 0.625 & 0.741 & 0.613 \\
\hspace{0.2in} - No random shuffling of questions & 0.515 & 0.313 & 0.523 & 0.319 \\\bottomrule
\end{tabular}
\vspace{0.5em}
\caption{Results of ablation studies on global re-ranker }
\label{tab:ablation}
\end{table}

\section{Discussion}
\label{sec:discussion}

In this work, we address an important problem of closing the training-inference gap for automated medical history-taking. Our approach, inspired by modern neural information retrieval systems, has two stages: (1) we use an expert system to suggest a list of candidate questions (out of possible thousands), (2) we train a machine-learned re-ranking model to re-rank expert system-suggested questions based on the free text of the doctor-patient dialogue.

To perform re-ranking (stage 2), we introduce a new approach which we call ``global re-ranker'', and compare it to existing neural baselines. We also explore several language model back-bones including various transformers and structure-state-space (S4) models\footnote{As part of this publication, we release bi-directional and autoregressive S4 checkpoints pre-trained on the English Wikipedia and Pubmed PMC Open Access Subset.}. We find that while all neural re-ranking models outperform the original expert system, the global re-ranker with a full-attention transformer backbone performs the best with a 30\% increase in nDCG and 77\% increase in mAP over the first-stage recommendations. 


While our results directly show the effectiveness of training a re-ranking model on top of an expert system for history taking, we believe this approach can also be applied to other decision support systems. The conditions under which this approach is beneficial are the following: (1) There exists a scoring system that has a training-inference gap (2) The space of possible predictions is very large, and as such would require a lot of data to machine-learn from scratch. One example beyond history-taking where we believe these conditions are satisfied is medical diagnosis prediction. There are many expert-system-derived diagnosis models, and training a diagnosis model from scratch can be difficult as the space of possible diagnosis is very large. Re-ranking could be used to close the gap between an off-the-shelf diagnostic expert system and the practice's actual patient population outcomes.

\paragraph{Limitations}
This work is still limited in several ways. While our proposed global re-ranker had exhibited best overall performance over other ranking models, it is still computationally inefficient as the full attention transformers have quadractic computational complexity in processing long sequence. This will become a more serious bottleneck as the dialogue gets longer or the number of candidate questions increases. Secondly, the global re-ranker only learns the association between history-taking questions and the dialogue contexts from languages but it does not have the underlying medical knowledge. It will be paramount to augment such models with real medical knowledge such that it makes more informed decisions and does not biased against low frequency long-tail history-taking questions. In the future, we plan to investigate more effective approaches to encode long textual contexts and inject knowledge into language mdoels.   

\subsection{Ethics} This work was done as part of a quality improvement activity as defined in 45CFR §46.104 (d)(4)(iii) -- secondary research for which consent is not required for the purposes of “health care operations.” 


\bibliographystyle{plainnat}

\bibliography{references}
\end{document}